\documentclass[wcp]{jmlr}

\usepackage{longtable}

\usepackage{booktabs}

\usepackage{multirow}
\usepackage{caption}
\usepackage{subcaption}

\jmlrvolume{101}
\jmlryear{2019}
\jmlrworkshop{ACML 2019}

\title[Variational Conditional GAN for Fine-grained Controllable Image Generation]{Variational Conditional GAN for Fine-grained Controllable Image Generation}

  \author{\Name{Mingqi Hu} \Email{mingq@seu.edu.cn}\\
  \addr Southeast University, Nanjing 211189, China
  \AND
  \Name{Deyu Zhou} \Email{d.zhou@seu.edu.cn}\\
  \addr Southeast University, Nanjing 211189, China
  \AND
  \Name{Yulan He} \Email{yulan.he@warwick.ac.uk}\\
  \addr University of Warwick, Coventry CV4 7AL, UK
 }
\editors{Wee Sun Lee and Taiji Suzuki}

\begin{document}

\maketitle

\begin{abstract}
In this paper, we propose a novel variational generator framework for conditional GANs to catch semantic details for improving the generation quality and diversity. Traditional generators in conditional GANs simply concatenate the conditional vector with the noise as the input representation, which is directly employed for upsampling operations. However, the hidden condition information is not fully exploited, especially when the input is a class label. Therefore, we introduce a variational inference into the generator to infer the posterior of latent variable only from the conditional input, which helps achieve a variable augmented representation for image generation. Qualitative and quantitative experimental results show that the proposed method outperforms the state-of-the-art approaches and achieves the realistic controllable images.
\end{abstract}

\section{Introduction}
Generating controllable images from natural language has many applications, including text illustration, computer-aided design and second-language learning. In recent years, it has gained increasing interests in the research community~\citep{mansimov2015generating}. However, due to the challenges faced in  language understanding and cross-modal transformation, it is far from being solved~\citep{reed2016generative}.

Since directly generating images from text descriptions is difficult, a relatively easier problem is to generate an image from a class label which indicates an image category such as `\emph{Plane}', `\emph{Cat}' and `\emph{Coat}', which is called \emph{class-conditional image generation}. Recently, with the emergence of Conditional Generative Adversarial Network (CGAN)~\citep{mirza2014conditional}, there has been remarkable progress in this field~\citep{denton2015deep,odena2016conditional,miyato2018cgans}. To the best of our knowledge, most generators in CGAN feed the class condition information (i.e., an class label vector), $c$, by simply concatenating $c$ with noise $\varphi$ to form an input which is then directly fed into the upsampling operations to generate an image. 
Another generative model, Conditional Variational Auto-Encoder (CVAE)~\citep{sohn2015learning}, also shows some promise for conditional image generation~\citep{mansimov2015generating,yan2016attribute2image}. However, in such approaches, the generated images are usually blurry because of direct sampling from prior and the element-wise measures used.

In this paper, inspired by the variation inference employed in VAE, we propose a novel variational generator framework for CGAN. More concretely, in our proposed framework, variational inference only depending on the conditional vector $c$ is introduced into the generator to infer the distribution of latent variable $z$, which represents the shared semantics across both text and image modalities. As the images are drawn from the inferred posterior in the generation phase, it overcomes the problem of mismatching from prior sampling in CVAE. Also, the reconstruction loss is augmented by minimizing the distance between the fake and true distributions through adversarial training. From another perspective, instead of upsampling from the concatenated representation in CGAN, the condition information is relatively fully exploited by posterior inference and a variable augmented representation (drawn from latent distribution) is supplied for image generation. By introducing variational inference, the fine-grained images with more visual details and richer diversity under the conditional constraint can be generated. The mode collapse problem that all outputs moving toward one or some fixed point can be partially addressed. 

The main contributions of this paper are summarized as follows:
\begin{itemize}
\item We propose a novel variational generator framework for CGAN to improve the generation quality and diversity. The framework is flexible and can be applied in various tasks such as text-to-image generation. To the best of our knowledge, it is the first attempt to incorporate the variational inference only from the conditional input (without images) into the generator of CGAN, which guarantees images can be generated from the posterior after training.

\item We propose a novel auxiliary classifier to better satisfy the class-conditional constraint. Experimental results show it accelerates adversarial training and avoid mode collapse problem than original version. We also incorporate a truncation technique as post-processing for latent space to further boost the generation performance.

\item Qualitative and quantitative experiments are conducted on the class-conditional task, and results show that the proposed method outperforms the state-of-the-art approaches for class-conditional image generation. We also applied our method to the text-to-image generation task, the fine-grained images that match the sentence descriptions can be achieved.
\end{itemize}

\section{Related work}
Research on image generation based on natural language can be classified into two categories: \emph{sentence-level image generation} and \emph{class-conditional image generation}. \emph{Sentence-level image generation} learns to generate related image from one sentence, which is also called text-to-image generation~\citep{reed2016generative}.  A number of end-to-end methods have been exploited to solve the problem.~\citet{mansimov2015generating} built an AlignDRAW model based on the recurrent variational auto-encoder to learn the alignment between text embeddings and the generating canvas. With CGAN,~\citet{reed2016generative} generated plausible images for birds and flowers based on text descriptions. Following this way,~\citet{hong2018inferring} generated more fine-grained images by decomposing the generation process into multiple steps. However, the text descriptions that are used for generation usually have simple grammatical structures only with single entity (e.g. ``\emph{This bird is red and brown in color, with a stubby beak.}"). Moreover, there is no effective quantitative metric to measure the consistency of the generated image and a given text description. The end-to-end methods also lack the interpretability.

\emph{Class-conditional image generation}~\citep{van2016conditional,odena2016conditional} is dedicated to generate images from an image class label (i.e., an entity). It is relatively easier compared to sentence-level image generation since it does not require the understanding of semantic meanings encoded in sentences. 
Some quantitative metrics such as Inception score (IS) and Frechet Inception distance (FID) have been exploited to evaluate the quality and variety of the generated images. Based on CVAE,~\citet{yan2016attribute2image} generated face images from the visual attributes extracted from a text description, through disentangling foreground and background of image. However a disadvantage of CVAE is that, the generated samples are often blurry because of the injected prior noise in the test phase and imperfect element-wise measures such as the squared error. Generative Adversarial Networks (GANs)~\citep{goodfellow2014generative} have shown promising performance for generating sharper images. Its variant, conditional GAN, has become a general framework for cross-modality transformation for various tasks (e.g., image-to-image translation and image captioning) by using conditional information for the discriminator and generator.~\citet{odena2016conditional} proposed a novel approach to incorporate the class-conditional information into the discriminator by adding an auxiliary classifier. Similarly,~\citet{miyato2018cgans} proposed a projection-based discriminator to further improve the generation quality. Instead of improving the discriminator.~\citet{zhang2018self} used the self-attention mechanism in the generator to make the generated image look more globally coherent.

However, to the best of our knowledge, most of methods based on CGAN generate images through a series of upsampling operations from the condition and noise. The concatenation representation of input is relatively simple and the hidden condition information is not fully exploited. It would be interesting to take into account the hidden semantics behind the conditional input to generate images. Recently, \citet{bao2017cvae} proposed a framework called CVAE-GAN, whcih uses the generator in CGAN as the decoder in CVAE to combine them and a feature matching loss to reconstruct images in feature space. However, it is still a CVAE-based framework, which can not overcome the blurry side effect by CVAE/VAE. Different from CVAE-GAN, we introduce a variational inference from control condition into the generator and the encoder can be reused in test phase for posterior inference, which we believe is a natural way to incorporate both advantages of VAE and GAN.

\section{Methodology}
We cast class-conditional image generation as a conditional likelihood maximization problem and define the problem setting in Section 3.1, followed by the proposed framework in Section 3.2. After that, we describe the training procedure in Section 3.3 and conditional data generation in Section 3.4.

\subsection{Problem Setting}
Given the condition variable $c \in \mathbb{R}^{N_c}$ (i.e., entities or image classes) and latent variable $z \in \mathbb{R}^{N_z}$, we aim to build a conditional generative model $p_\theta(x|z)$ to generate a realistic image $x \in \mathbb{R}^{N_x}$ conditioned on $z$, which is the hidden semantics behind $c$.

A traditional way of model learning is to maximize the variational lower bound of the conditional log-likelihood $\log p_\theta(x|c)$. Specifically, an auxiliary distribution $q_\phi(z|x, c)$ is introduced to approximate the true posterior $p_\theta(z|x, c)$. The conditional log-likelihood can be formulated below~\citep{yan2016attribute2image}:
\vspace{-0.1in}
\[ \begin{split}
\log p_\theta (x|c) &= KL(q_\phi (z|x, c)||p_\theta (z|x,c)) 
+ \mathcal{L}(x,c;\theta,\phi),
\end{split} \]
\vspace{-0.3in}
\begin{equation} \begin{split}
\mathcal{L}(x,c;\theta,\phi) = -KL( q_\phi (z|x, c)||p_\theta (z)) 
- \mathbb{E}_{q_\phi(z|x,c)}L(g_\theta(c, z), x)
\end{split} \end{equation}
where the Kullback-Leibler divergence $KL(q_\phi(z|x, c)||p_\theta(z))$ as a regularization loss reduces the gap between the prior $p_\theta(z)$ and the auxiliary posterior $q_\phi(z|x, c)$, and $L(g_\theta(c,z),x)$ is the reconstruction loss (e.g., $\ell_2$ loss) between the generated image $g_\theta(c, z)$ and real image $x$. An encoder network $f(x,c)$ and a decoder network $g(c,z)$ are built for $q_\phi(z|x, c)$ and $p_\theta(x|c, z)$, respectively. During training, image is generated based on $g(c,z)$ where $z$ is drawn from the inferred posterior $q_\phi(z|x,c)$. However, in the test phase, latent samples $z$ are all drawn from the prior distribution $p_\theta(z)$ (usually, $z$ is a Gaussian noise) instead of the posterior distribution as the real image $x$ is not given. Obviously, there is a mismatch between the latent sample $z \sim p_\theta(z)$ and the condition $c$, which might generate unclear images. This mismatch problem is inherent as the latent distribution is not modeled explicitly.

We can however assume that the latent space has the local aggregated property, depending on the different conditions (manifold hypothesis). Based on the assumption, we directly infer the latent variable posterior $p_\theta(z|c)$ only depending on condition $c$ (without image $x$) so that the images can be drawn from the inferred posterior $q_\phi(z|c)$ instead of the prior. Such modification brings another problem: there is no ground-truth $x_{c_i}$ to calculate reconstruction loss $L(g_\theta(z), x_{c_i})$ directly as a specific condition $c_i$ usually corresponds to a variety of real images $x_{c_i}$. Thus an adversarial loss $- \log D(g_\theta(z)|x, c)$ given by a discriminator $D$ is introduced as the reconstruction loss to help reduce the distance of the generated images and the ground-truth. The total objective function of the encoder-decoder network is modified as below:
\vspace{-0.1in}
\begin{equation} \begin{split}
 \mathcal{L}(x,c, \varphi; ~\theta,\phi) = -KL( q_\phi (z|c, \varphi)||p_\theta (z)) 
+ \mathbb{E}_{q_\phi(z|c, \varphi)}\log D(g_\theta(z) | x, c)
\label{cvae2}
\end{split} \end{equation}
where noise $\varphi$ is introduced as a part of the input of the encoder to compensate the disturbance caused by $x$.

\vspace{-0.1in}
\subsection{The Proposed Framework}

To solve the objective function defined in ${\rm Eq.~\ref{cvae2}}$, we design a novel model architecture named Variational Conditional GAN (VCGAN). The VCGAN architecture as shown in {\rm Figure~\ref{vcgan_framework}} contains the following three modules: (1) \emph{Encoder Network} $F_{\phi}(c ,\varphi)$, takes noise $\varphi$ and condition $c$ as input and encodes them as latent variable $z$; (2) \emph{Decoder Network} $G_{\theta}(z)$, is designed to learn the distribution of real image $x$ given the latent variable $z$; (3) \emph{Discriminator Network} $D(x)$, as a supervisor to judge the image $x$ and supply the reconstruction loss for the \emph{encoder-decoder network} $G(c, \varphi)$ (the proposed variational generator framework).
We explain the structure and function of each module of VCGAN in more details below:
\begin{figure*}[!htbp]
\centering
\vspace*{-0.3in}
\includegraphics[scale=0.6]{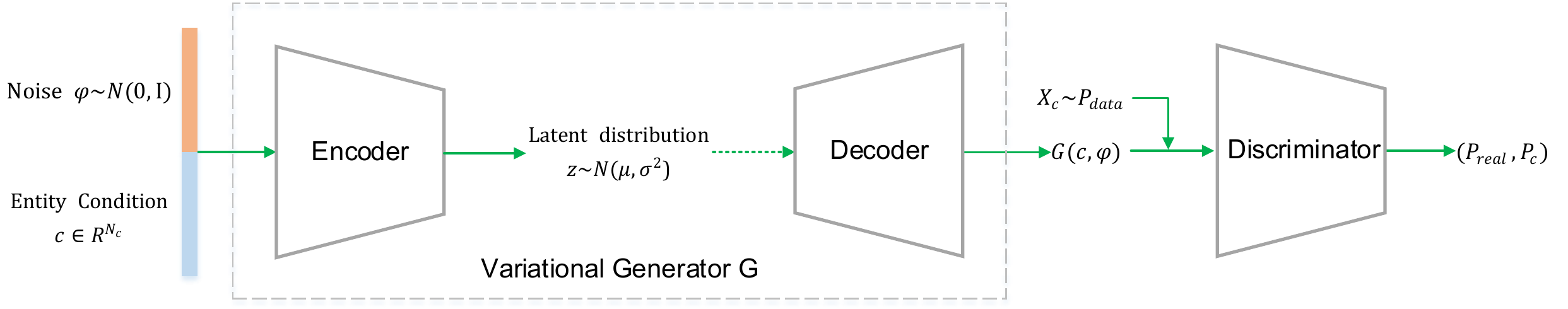}
\caption{The Architecture of the Variational Conditonal GAN (VCGAN). The dashed line represents the sampling operation.}
\vspace*{-0.2in}
\label{vcgan_framework}
\end{figure*}

\noindent \textbf{Encoder} $F_{\phi}(c, \varphi)$: The encoder aims to conduct posterior inference of latent variable $z$ given the condition variable $c \in \mathbb{R}^{N_c}$ with noise variable $\varphi \sim \mathcal{N}(0, I)$. The posterior $q_{\phi}(z|c,  \varphi)$ is assumed as a diagonal Gaussian where the mean and covariance are parametrized by encoder $F_{\phi}(c, \varphi)$.
\vspace{-.15in}
\begin{equation} \begin{split}
&\varphi \sim \mathcal{N}(0, I) \\
&(\mu, diag(\sigma^2)) = F_{\phi}(c, \varphi) \\
&z \sim \mathcal{N}(\mu, diag(\sigma^2))
\end{split} 
\vspace{-0.3in}
\end{equation}
where noise $\varphi$ is drawn from standard multivariate Gaussian, the mean $\mu$ and covariance $diag(\sigma^2)$ of the latent variable $z$ are estimated by the encoder.

\begin{figure}[!h]
\centering
\vspace*{-0.2in}
\includegraphics[scale=0.7]{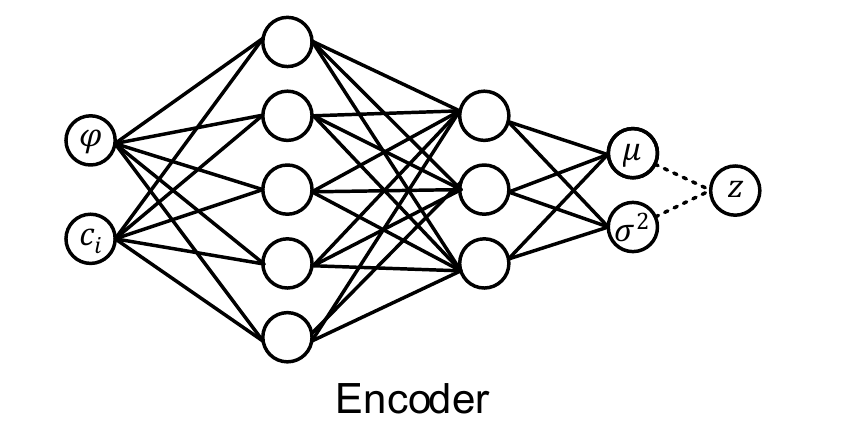}
\vspace*{-0.15in}
\caption{The architecture of the Encoder in VCGAN. Each sample is drawn from the latent distribution by the \emph{reparameterization trick}.}
\vspace*{-0.1in}
\label{encoder}
\end{figure}

The structure detail of the encoder is shown in {\rm Figure~\ref{encoder}}, we employed three linear layers in our experiments for simplicity. More complex neural network architecture could be used in the encoder for inference.

\vspace{-0.1in}
\noindent \textbf{Decoder} $G_{\theta}(z)$: The decoder learns a function $f: \mathbb{Z} \rightarrow \mathbb{X}$, where $\mathbb{Z}$ represents the low dimensional latent space and $\mathbb{X}$ represents the high dimensional pixel space. The decoder takes the latent variable $z$ drawn from the posterior $q_{\phi}(z|c, \varphi)$ as input and decodes it as the image $x$. The decoding process can be formulated below:
\vspace{-.1in}
\begin{equation} \begin{split}
&z \sim q_{\phi}(z|c, \varphi)~;~x = G_{\theta}(z)
\end{split} \end{equation}

\vspace{-0.1in}
\noindent \textbf{Discriminator} $D(x)$: To calculate the reconstruction loss, we introduce the discriminator as a supervisor to reduce the distance of the generated images and the ideal ground-truth. Specifically it judges the perceptual fidelity of the generated image $x$ and the consistency with corresponding condition $c$. In order to satisfy the above two requirements, we design a novel objective based on a auxiliary classifier~\citep{odena2016conditional}. The discriminator outputs both a probability distribution over sources $s$, $p(s | x)$, and a probability distribution over the conditions $c$, $p(c | x)$. The objective function of $D(x)$ includes two parts: the log-likelihood of the correct source, $L_s$, and the log-likelihood of the correct condition, $L_c$.
\vspace{-0.1in}
\begin{equation} \begin{split}
L_{s} = \mathbb{E} [ \log p ( s = real | x _ { real } ) ] + \mathbb{E} [ \log p ( s = fake | x _ { fake } ) ]
\label{ns-gan}
\end{split} \end{equation}
\vspace{-0.25in}
\begin{equation} \begin{split}
L_{c} = \mathbb{E} [ \log p ( c = c_x | x _ { real } ) ] + \mathbb{E} [ \log p ( c = Others | x _ { fake } ) ]
\label{other}
\end{split} \end{equation}
where ${\rm Eq.~\ref{ns-gan}}$ is the standard GAN objective~\citep{goodfellow2014generative}, it minimizes the negative log-likelihood for the binary classification task (\emph{is the sample true or fake?}) and is equivalent to minimize the Jensen-Shannon divergence between the true data distribution $P$ and the model distribution Q.

It should be pointed out that we introduce $\mathbb{E} [ \log p ( c = Others | x _ { fake } ) ]$ in ${\rm Eq.~\ref{other}}$ instead of the term $\mathbb{E} [ \log p ( c = c_x | x _ { fake } ) ]$ as in original version. 
This is to introduce an additional class ``\emph{Others}" to represent that the image $x$ does not belong to any of the known conditions. Experimental results show that labeling the $x_{fake}$ with ``\emph{Others}" makes the training converge faster. 

We inline the modified classifier to the discriminator as a one-ve-many multi-task learning by sharing the hidden layers to improve each other. Suppose there are a total of $K$ classes or conditions, then we have $K+2$ output units in the discriminator network. We use a Sigmoid activation function in the first output unit and the Softmax function over the remaining K+1 output units (include an extra class ``\emph{Others}"). The probability of the image $x$ belonging to sources $s$ and conditions $c$ can be written below:
\begin{flalign} \begin{split}
&P ( s = real | x ) = sigmoid\left( h(x) \cdot \boldsymbol { W }_{0} + b _ { 0 } \right) \\
&P ( c = c_i | x ) = \frac { \exp \left( h(x) \cdot \boldsymbol { W }_{i} + b _ { i } \right) } { \sum _ { i = 1 } ^ { K+1 } \exp \left( h(x) \cdot \boldsymbol { W } _ { i } + b _ { i } \right) }
\end{split} \end{flalign}
where $h(x)$ is the hidden layer representation of image $x$, $\boldsymbol{W} \in \mathbb{R}^{H\times (K+2)}$ denotes the weight matrix of the output layer, $\boldsymbol{ W }_0$ is the weights of the first output unit. 

\subsection{Training}
\label{training}
The final loss functions of the proposed framework are given as follows:
\begin{equation} \begin{split}
\mathcal{L}_{D} = & -( \mathbb{E}_{x \sim P} [ \log D(x)_0 ] + \mathbb{E}_{x \sim Q} [ \log(1- D(x)_0) ] ) \\
			       & - (\mathbb{E}_{x \sim P}[\log D(x)_1] + \mathbb{E}_{x \sim Q}[\log D(x)_1])
\end{split} \end{equation}
\begin{equation} \begin{split}
\mathcal{L}_{G} = &- \mathbb{E}_{x \sim Q} [ \log D(x)_0 ] - \mathbb{E}_{x \sim Q}[\log D(x)_1] 
+ KL( q(z|c, \varphi)||p(z))
\label{Lg}
\end{split} \end{equation}
where $P$ is true data distribution and Q is the generated data distribution. The $D(x)_0$ denotes the probability of image $x$  being real, the $D(x)_1$ denotes the the probability over the label condition (including the ``\emph{Others}" class). 
The KL divergence of the latent prior $p(z)$ from $q(z|c, \varphi)$ is added to $\mathcal{L}_{G}$ in ${\rm Eq.~\ref{Lg}}$, as the regularization loss for constraining the latent $z$.

Based on the assumptions that the latent posterior is a diagonal Gaussian with mean $\mu$ and standard deviation $\sigma$ and the prior is a standard Gaussian with mean 0 and standard deviation 1, in which case the $KL$ term in ${\rm Eq.~\ref{Lg}}$ becomes
\vspace{-.1in}
\begin{equation} \begin{split}
KL( q(z|c, \varphi)||p(z))=  -\frac { 1 } { 2 } \sum_{ j = 1 } ^ { J } \left( 1 + \log (( \sigma _ { j }) ^ { 2 }) - ( \mu _ { j } ) ^ { 2 } - ( \sigma _ { j  } ) ^ { 2 } \right)
\end{split} \end{equation}

In the training stage, we first optimize the discriminator $D(x)$ under the loss $\mathcal{L}_{D}$ with the fixed generator, and then optimize the generator $G(c,\varphi)$ under the loss $\mathcal{L}_{G}$ (note the generated image is labeled with the fed condition $c$ not ``Others'' when training G) with the fixed discriminator. The above two steps alternates (adversarial training) by minibatch stochastic gradient descent.

As GAN training is relatively unstable and hard to optimize, we investigate some techniques to stabilize the adversarial training. We use batch normalizationin both the generator and the discriminator, and only use spectral normalization~\citep{miyato2018spectral} in the discriminator to make the training more stable.

\subsection{Conditional Image Generation}
An image $x$ can be generated controllably from a learned VCGAN model by picking latent sample $z$ from a inferred posterior $q(z|c, \varphi)$, then running the decoder to generate the image $x$ fulfilling the condition $c$. The generation process in test phase can be described in the following:
\vspace{-.1in}
\begin{equation} \begin{split}
&\varphi \sim \mathcal{N}(0, I) \\
&q(z|\varphi, c) = F_{\phi}(c, \varphi) \\
&z \sim q(z|c, \varphi) \\
&x = G_{\theta}(z)
\end{split} \end{equation}
Note that we reuse the encoder to obtain inferred posterior $q(z|\varphi, c)$ not the fixed prior $p(z)$ as in CVAE. 

We also incorporate a \textit{Truncation Technique} as a post-processing step for further enhancing the quality of the generated images. Specifically, we draw the latent sample $z$ from a truncated posterior (e.g., a truncated Gaussian distribution) then feed it into the decoder. 
The latent variable $z$ will be resampled if it exceeds the truncation range. Using this technique, we can achieve a higher generation quality than sampling from the whole latent space.

\vspace{-.05in}
\section{Experiments}

We first conduct extensive class-conditional experiments on two popular datasets, CIFAR10~\citep{krizhevsky2009learning} and FASHION-MNIST~\citep{xiao2017fashion}, which are widely employed by many state-of-the-art conditional image generation approaches. The two datasets both include ten different categories of objects. Also, we conduct sentence-level experiments on another two datasets, CUB~\citep{wah2011caltech} and Oxford Flower~\citep{nilsback2008automated}, which include 10 captions for each image provided by \citet{reed2016representation}. They include many subcategories of birds and flowers. The statistics of the these datasets are presented in Table \ref{datasets}.
\begin{table}[!htbp]
\centering
\resizebox{\columnwidth}{!}{
\begin{tabular}{l|cccc}
\hline
Datasets        & CIFAR10          & FASHION-MNIST           &CUB                     &Oxford Flower                                                      \\ 
\hline
Training images & 50,000          & 60,000                              &8,855                    &7,034                                                                                   \\
Test images     & 10,000            & 10,000                              &2,933                    &1,155                                                                                \\
Resolution       & 32x32              & 28x28                               &128x128               &128x128                                                                                                        \\
Class labels     & \multicolumn{1}{l}{\begin{tabular}[c]{@{}l@{}}``Plane",``Car",``Bird",\\ ``Cat",``Deer",``Dog",``Frog",\\ ``Horse",``Ship",``Truck"\end{tabular}}    & \multicolumn{1}{l}{\begin{tabular}[c]{@{}l@{}}``Tshirt",``Trouser",``Pullover",\\ ``Dress",``Coat",``Sandal",``Shirt",\\ ``Sneaker",``Bag",``Ankle boot"\end{tabular}}    & 200   & 102 
\\ 
Captions          & \textbf{--}                       & \textbf{--}                                          & 10                         & 10       
\\
\hline
\end{tabular} }
\caption{Statistics of the datasets.}
\vspace*{-0.1in}
\label{datasets}
\end{table}

Two quantitative metrics, Inception score (IS)~\citep{salimans2016improved} and Frechet Inception distance (FID)~\citep{heusel2017gans} are employed. IS uses an pre-trained Inception net on ImageNet to calculate the statistic of the generated images, which is defined as:
\vspace*{-0.1in}
\begin{equation} \begin{split}
{\rm{IS}} = \exp\left(\mathbb{E}_{x \sim Q} [ KL (p(y|x) || p(y) ) ] \right)
\label{IS}
\end{split} \end{equation}
where $p(y|x)$ is the conditional class distribution given by Inception net, and $p(y) = \int_x p(y|x)p(x)$ is the marginal class distribution. The higher IS indicates the generated images contain clear recognizable objects. However, as pointed out in~\citep{zhang2018self}, IS can not assess the perceptual fidelity of details and intra-class diversity. FID is a more principled metric, which can detect intra-class mode collapse~\citep{kurach2018gan}. By assuming that the sample embeddings follow a multivariate Gaussian distribution, FID measures the Wasserstein-2 distance between two Gaussian distributions, which is defined as:
\vspace*{-0.2in}
\begin{equation} \begin{split}
\mathrm { FID } = \left\| \mu _ { x_1 } - \mu _ { x_2 } \right\| _ { 2 } ^ { 2 } + \operatorname { Tr } \left( \Sigma _ { x _1} + \Sigma _ { x_2 } - 2 \left( \Sigma _ { x_1 } \Sigma _ { x_2 } \right) ^ { 1/2 } \right)
\label{FID}
\end{split} \end{equation}
where $x_1$ and $x_2$ are samples from $P$ and $Q$. The lower FID value, the closer distance between the synthetic and the real data distributions. In all our experiments, 50k samples divided into 10 groups are randomly generated to compute the Inception scores and 10k samples are generated to compute FID.

Seven state-of-the-art approaches chosen as the baselines:
\begin{itemize}
\item DCGAN~\citep{radford2015unsupervised}, a deep convolutional architecture of GAN.
\item LSGAN~\citep{mao2017least}, which uses a least-squares loss instead of the standard GAN loss which minimizes the Pearson $\chi^2$ divergence between $P$ and $Q$.
\item AC-GAN~\citep{odena2016conditional}, an auxiliary classifier is introduced into the discriminator for class-constraint.
\item WGAN~\citep{arjovsky2017wasserstein}, which minimizes the Wasserstein distance between $P$ and $Q$.
\item WGAN-GP~\citep{gulrajani2017improved}, which uses the Gradient Penalty as a soft penalty for the violation of 1-Lipschitzness in WGAN.
\item CVAE-GAN~\citep{bao2017cvae}, a CAVE-based framework with a discriminator for feature matching.
\item SNHGAN-Proj~\citep{miyato2018cgans}, a spectrally normalized GAN was combined with the projection-based discriminator with hinge loss.
\end{itemize}

The proposed model was implemented based on a recent robust architecture called SNDCGAN~\citep{miyato2018spectral}. Another common architecture is ``ResNet''~\citep{gulrajani2017improved}. It is deeper than SNDCGAN, which we use in the sentence-level experiments. The proposed encoder is implemented by three-layer FCs with 512, 256, and 128 units. The CVAE-GAN is also implemented by SNDCGAN with a convolutional image encoder for better comparsion with ours. For all experiments, we fix the size of the latent variable and noise variable to 128, encode the class conditions as one-hot representations. We choose Adam solver with hyper-parameters set to $\beta_1 = 0.5, \beta_2= 0.999$ and the learning rate $\alpha = 0.0002$ by default. The batch size is set to 100 and the balanced update frequencies (1:1) of discriminator and generator are employed. For sentence-level experiments, a smaller batch size (64) are employed. We use the char-CNN-RNN text encoder provided by \citet{reed2016representation} to encode each sentence into a 1024-d text embedding as the conditional vector.
\vspace*{-0.25in}
\begin{figure}[!htbp]
\centering
	\begin{minipage}[b]{.4\textwidth}
	\centering
	\includegraphics[width=\textwidth]{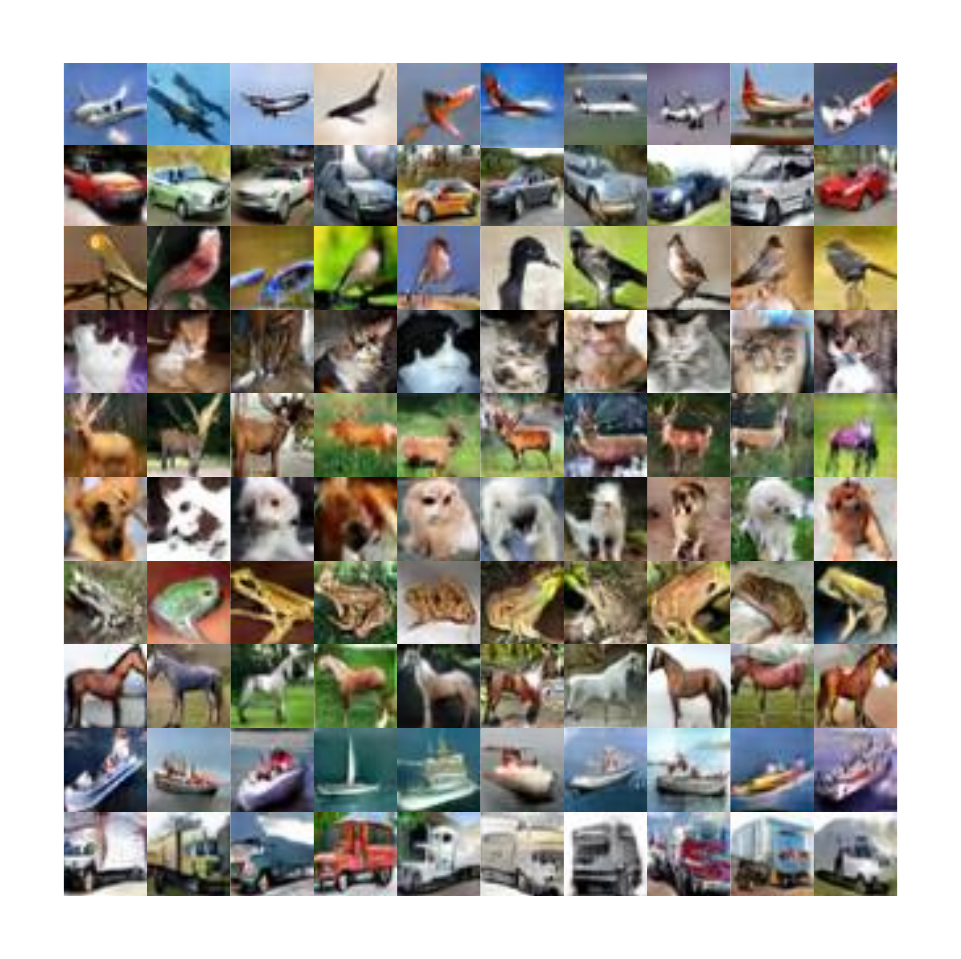}
	\subcaption{CIFAR10 samples.}
	\label{cifar10-fig}
	\end{minipage}
	\begin{minipage}[b]{.4\textwidth}
	\centering
	\includegraphics[width=\textwidth]{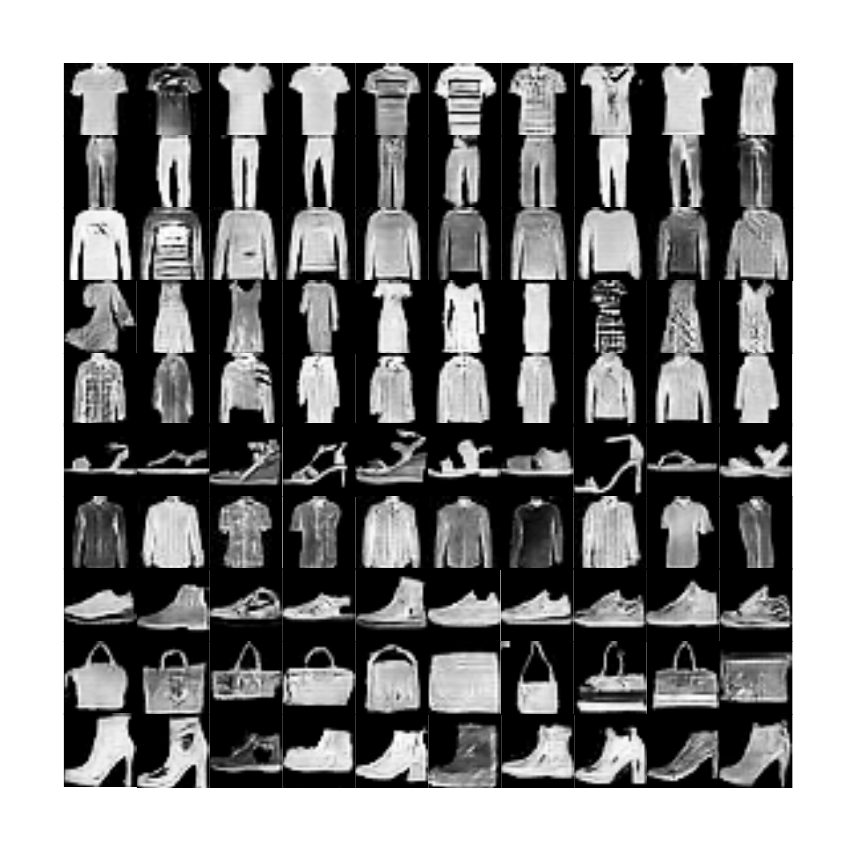}
	\subcaption{FASHION-MNIST samples.}
	\label{fashion-fig}
	\end{minipage}
\label{cifar-fashion-show}
\vspace*{-0.1in}
\caption{The generated samples controlled by the class labels with the proposed model.}
\end{figure}

\vspace{-.25in}
\subsection{Performance Comparison}

To evaluate the effectiveness of the proposed model, the class-conditional experiments are conducted on the CIFAR10 and FASHION-MNIST datasets. The results are presented in the Table \ref{cifar} and Table \ref{fashion}. Sample images generated by VCGAN are illustrated in Figure~\ref{cifar10-fig} and \ref{fashion-fig}.

\begin{table}[!htbp]
\center
\begin{tabular}{l|ll}
\hline
Method                 & \multicolumn{1}{c}{Inception Score $\uparrow$ } & \multicolumn{1}{c}{FID $\downarrow$} \\ \hline
DCGAN				        & $6.58 $              & --                        \\
AC-GAN           		    	& $8.25 \pm .07$ & --                        \\
WGAN-GP (ResNet)       	& $8.42 \pm .10$ & 19.5                 \\
SNHGAN-Proj (ResNet)   	& $8.62 $              & 17.5                 \\
CVAE-GAN				& $7.92 \pm .10$ & 24.1                 \\
VCGAN       	& $\mathbf{8.90 \pm .16}$   & $\mathbf{16.9}$                \\ \hline
\end{tabular}
\caption{Comparison with the baselines on CIFAR10 (with truncation post-processing). Some results are collected from \citet{gulrajani2017improved} and \citet{miyato2018cgans}. }
\label{cifar}
\end{table}

\begin{table}[!htbp]
\center
\begin{tabular}{l|ll}
\hline
Method~~~~~~~~~~~~~~~~~~~~~~~~~~~~ & \multicolumn{1}{c}{Inception Score $\uparrow$} & \multicolumn{1}{c}{FID $\downarrow$} \\ \hline
DCGAN              & $4.15 \pm .04$	 & --                       \\
WGAN                & $3.00 \pm .03$ 	 & 21.5                \\
LSGAN               & $4.45 \pm .03$ 	 & 30.7                \\
CVAE-GAN	     & $4.28 \pm .04$      & 15.9                \\
VCGAN               & $\mathbf{4.75 \pm .06}$   & $\mathbf{13.8}$             \\ \hline
\end{tabular}
\caption{Comparison with the baselines on FASHION-MNIST. Some results are collected from \citet{nandy2018normal} and \citet{lucic2018gans}. }
\label{fashion}
\end{table}

It can be observed from Table~\ref{cifar} and \ref{fashion}, the proposed approach (VCGAN) achieves the best IS and FID on both datasets. On FASHION-MNIST, FID is significantly improved from 15.9 to 13.8 by VCGAN. Furthermore, some baselines relied on a deep ResNet network, which is more complex than SNDCGAN we used and needs more training resources. It should be pointed out that instead of carefully selecting some modified versions of GAN loss function (e.g., Wasserstein loss or Hinge loss) in the baselines, our VCGAN simply used the original standard GAN loss. And also, VCGAN significantly outperforms the similar model, CVAE-GAN, on two datasets with similar settings. Figure~\ref{cifar10-fig} and Figure~\ref{fashion-fig} shows the clear and realistc images with hige variety can be achieved by VCGAN.

\begin{table*}[!htbp]
\centering
\resizebox{\textwidth}{!}{
\begin{tabular}{l|lllll|rrrrr}
\hline
Method                        & \multicolumn{5}{c|}{Inception Score $\uparrow$} & \multicolumn{5}{c}{FID $\downarrow$} \\ \hline
\multicolumn{1}{l|}{Truncation} & normal   & $2\sigma$   & $1.5\sigma$   & $\sigma$  & $0.5\sigma$  & normal  & $2\sigma$  & $1.5\sigma$  & $\sigma$  & $0.5\sigma$  \\ \hline
CVAE                          & $4.00 \pm .03$ & $4.12 \pm .05$ & $4.23 \pm .05$ & $4.52 \pm .05$ & $5.06  \pm .02$   & 105.9  & 103.5  & 100.7  & 95.5 & 91.9 \\
Concat-CGAN            & $6.51 \pm .08$ & $6.79 \pm .07$ & $6.98 \pm .08$ & $7.37 \pm .10$ & $7.08 \pm .05$   & 34.7 & 32.3  & 30.6 & 30.4 & 45.9 \\
CBN-CGAN                 & $7.33 \pm .10$ & $7.35 \pm .10$ & $7.33 \pm .08$ & $7.46 \pm .07$ & $7.29 \pm .09$   & 29.4 & 29.5  & 28.7 & 28.4 & $\mathbf{30.0}$ \\
CVAE-GAN                 & $7.72 \pm .08$ & $7.86 \pm .07$ & $7.92 \pm .10$ & $7.91 \pm .09$ & $7.37 \pm .08$   & 25.5 & 24.3 &24.1 & 28.0 & 45.6 \\
VCGAN                        & $\mathbf{8.43 \pm .13}$ & $\mathbf{8.62 \pm .10}$ & $\mathbf{8.90 \pm .16}$ & $\mathbf{8.80 \pm .13}$ & $\mathbf{7.66 \pm 0.05}$  & $\mathbf{17.6}$ & $\mathbf{16.9}$ & $\mathbf{17.3}$ & $\mathbf{20.5}$ & $37.2$ \\ \hline
\end{tabular}}
\caption{Ablation on CIFAR10 with different truncated ranges.}
\vspace*{-0.2in}
\label{ablation}
\end{table*}

\subsection{Ablation Comparison on CIFAR10}

To further evaluate the effectiveness of the proposed variational generator framework, we conduct an ablation comparison with CVAE and CGAN models on CIFAR10. We also make a comparison with CVAE-GAN. For CVAE, we follow the general setting by concatenating image and class condition as input and feeding it into the same architecture (without discriminator). For CGAN, there are two ways of feeding the condition $c$ into the generator (without encoder). There is usually to simply concatenate $c$ with the noise $\varphi$ as input as stated before, which we called Concat-CGAN. Recently,~\citet{miyato2018cgans} used conditional batch normalization~\citep{dumoulin2017learned} instead of batch normalization to feed condition information into the each layer of the generator (except the output layer), which we called CBN-CGAN. The same loss functions in section \ref{training} except $KL$ term are employed for Concat-CGAN and CBN-CGAN. The truncation technique mentioned is applied on all models and the truncated ranges are set by three-sigma rule, e.g., the range $2\sigma$ means samples are sampled (or re-sampled) from $\mu\pm2\sigma$, $normal$ means no truncation for the original latent space (noise space).

The results of various approaches on CIFAR10 are reported in Table~\ref{ablation}. It can be observed that our VCGAN significantly outperforms all opponents on different metrics and ranges. It shows that VCGAN has a more powerful generation capability and can perform latent posterior inference only from the class condition to achieve richer semantic details. Compared with CGAN, VCGAN achieves much lower FID values. It shows that by incorporating the variational inference into the generator, VCGAN can better approximate the real image distribution and owns richer diversity of images. Compared with CVAE-GAN, VCGAN is simple in framework and losses but more effective in results, which we attribute to the CGAN-based framework and the reusable encoder for posterior inference.

\begin{figure}[!htbp]
\centering
\vspace*{-0.1in}
\includegraphics[scale=0.8]{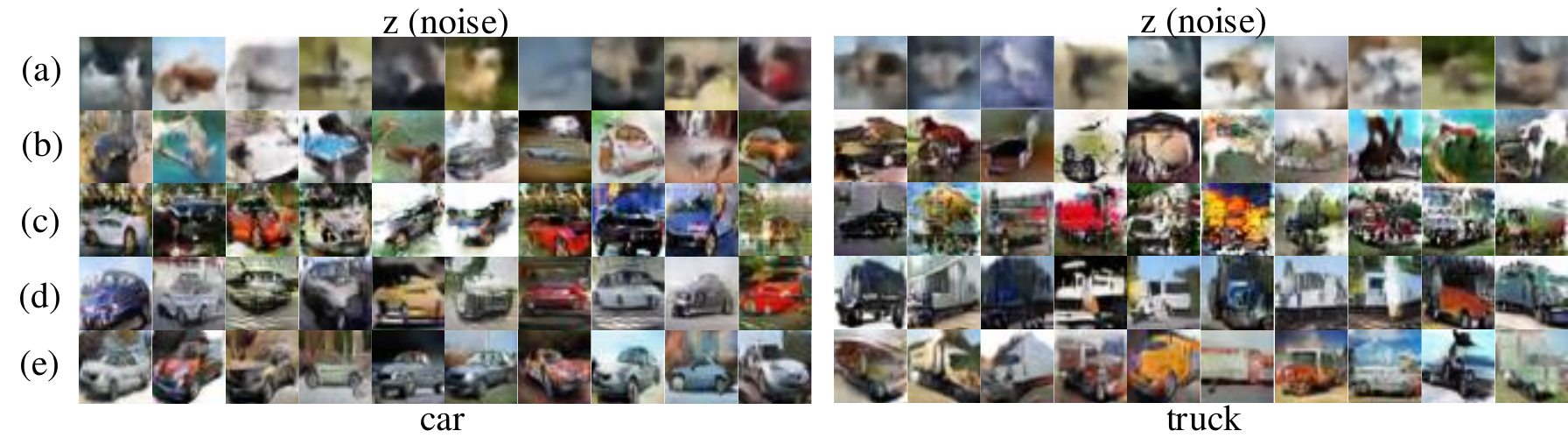}
\caption{Comparison of the images generated by (a) CVAE; (b) Concat-CGAN; (c) CBN-CGAN; (d) CVAE-GAN; (e) VCGAN. Each column corresponds to a fixed latent sample $z$ or noise.}
\label{ablation-imgs}
\end{figure}

Figure \ref{ablation-imgs} shows the samples generated by the five different approaches from various $z$ (or noise). The samples generated by CVAE seem very blurry even not related with the class. As stated before, CVAE suffers from the pixel-wise measure and the prior sampling. The Concat-CGAN and CBN-CGAN are more shaper, but there are much noise in the samples. Relatively, the CBN-CGAN performs better than Concat-CGAN.  CVAE-GAN is much better compared with the former, however it is blurry and distorted in the detail. Our proposed VCGAN model achieves clearer and more natural images with more visual details and also does well in the diversity.

\vspace{-.2in}
\subsection{Effect of Modified Auxiliary Classifier}
\begin{figure}[!h]
\vspace*{-0.1in}
\centering
	\begin{minipage}[b]{0.4\textwidth}
	\centering
	\includegraphics[width=\textwidth]{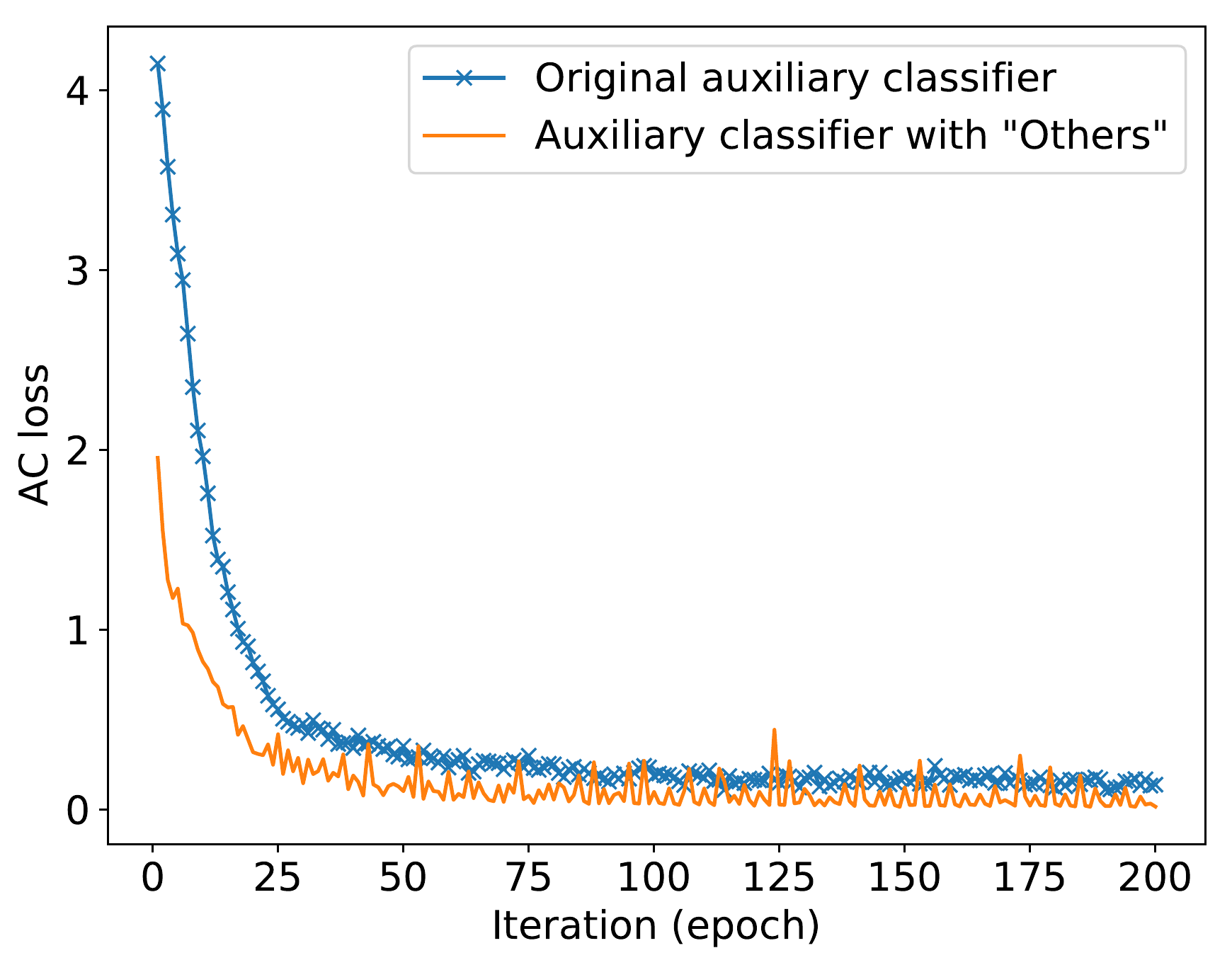}
	\subcaption{Training curves of the modified auxiliary classifier and the original version under the same setting.}
	\label{ac_loss}
	\end{minipage}
	\begin{minipage}[b]{0.4\textwidth}
	\centering
	\includegraphics[width=\textwidth]{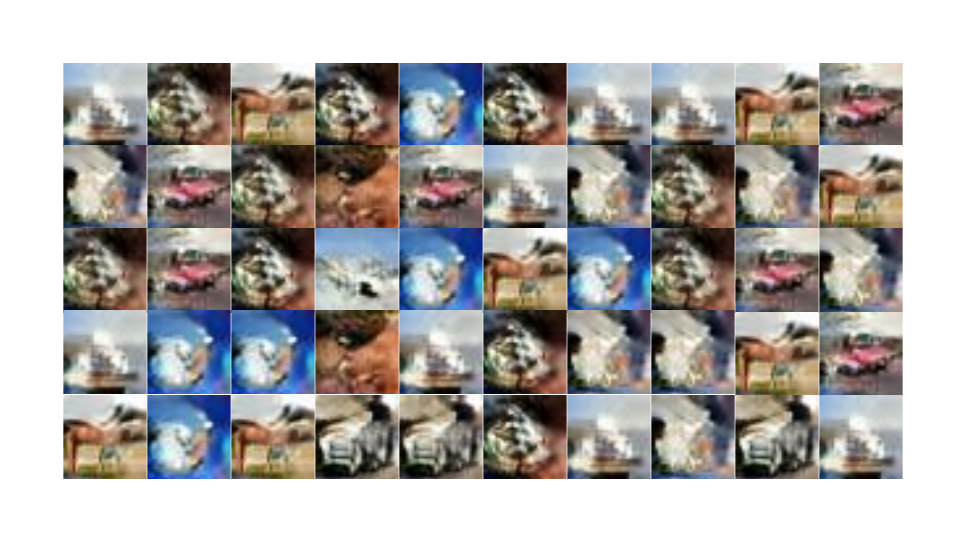}
	\subcaption{The collapsed images generated by original auxiliary classifier under random class-conditional input.}
	\label{dropping}
	\end{minipage}
\caption{The analysis for the modified auxiliary classifier on CIFAR10.}
\end{figure}

We further investigate the effect of incorporating the modified auxiliary classifier (AC) loss on CIFAR10. The training curves are shown as Figure~\ref{ac_loss}. It can be observed that the modified version coverages faster and has lower classification error, which helps the discriminator accelerate training and get closer to the optimal state. It can also be found that the images generated by the original AC loss might lead to the mode collapse~\citep{miyato2018cgans}, which is hardly discovered in the experiments using the modified AC loss. Some collapsed images by original auxiliary classifier are shown in Figure~\ref{dropping}.

\subsection{Sentence-level Image Generation} 

Although we believe it is relatively hard to generate an image from a sentence without the understanding for entity concepts, we also directly apply the proposed generator framework to the text-to-image generation task to evaluate the generalizability of the framework. Unlike class-conditional task, a text encoder is needed to embed the sentence semantics into the conditional vector. We use a hybrid character-level convolutional-recurrent network~\citep{reed2016representation} to calculate the semantic embeddings from text descriptions. The dimension of the sentence embedding is 1024. The dimension of noise and latent variable is 128. The output dimension of generator is set to 128 x 128. We conduct a pre-processing on CUB and Oxford Flower datasets, and all images are randomly cropped to 128 x 128 and flipped horizontally for data augmentation. Following the setup in~\citep{reed2016generative}, we split CUB and  Oxford Flower into class-disjoint training and test sets. we randomly pick an view (e.g., crop, flip) of the image and one of the captions as a pair for mini-batch training. The minibatch size is set to 64 and the model is trained for 300 epochs.

\begin{figure}[!htbp]
\centering
\vspace*{-0.2in}
\includegraphics[width= \textwidth]{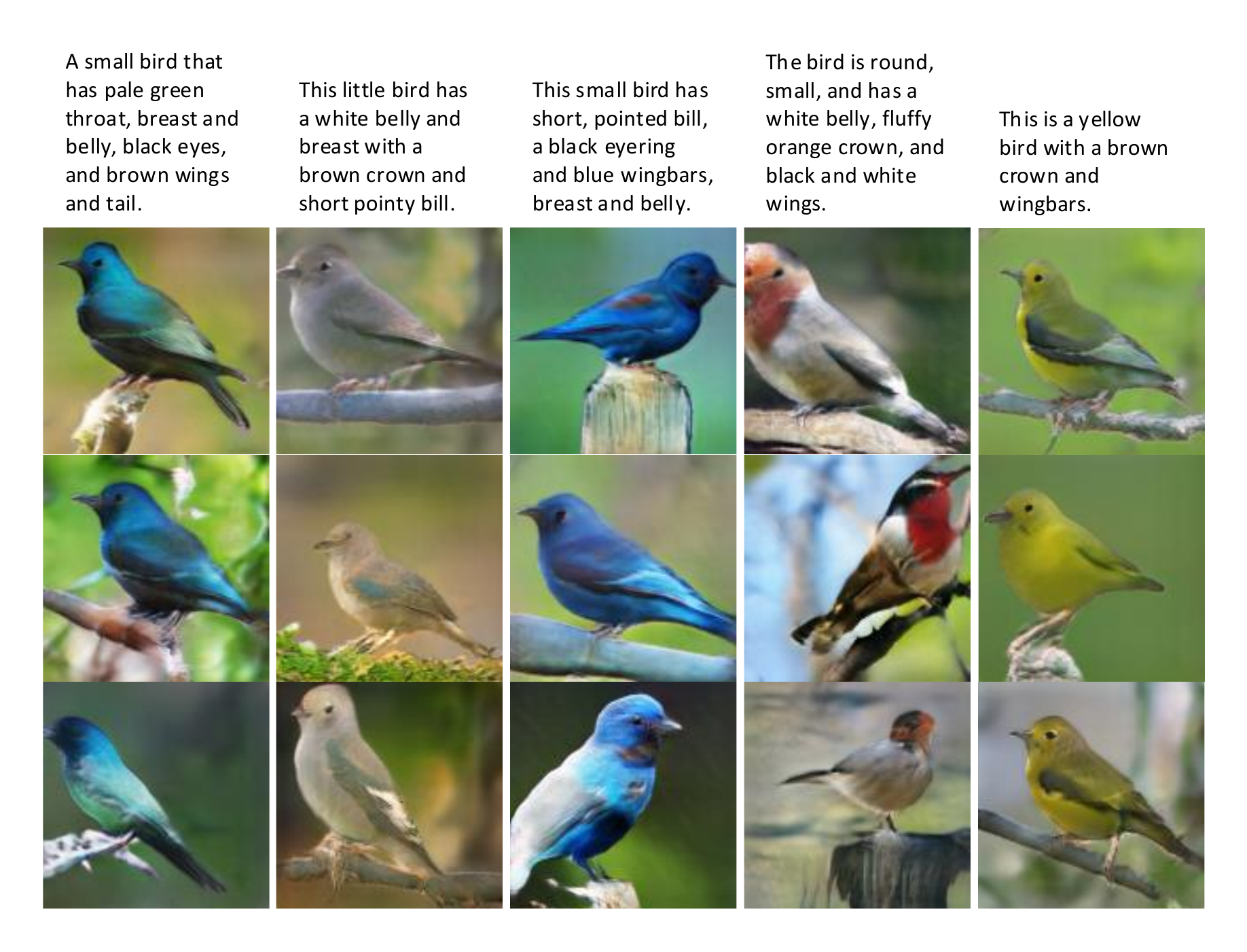}
\vspace*{-0.4in}
\caption{The generated 128 x 128 images controlled by text descriptions on CUB testset.}
\label{cub-sen}
\end{figure}

The Inception Scores on CUB is $5.00 \pm .11$ and on Oxford Flower is $2.95 \pm .03$. The generated samples controlled by text descriptions are presented in Figure~\ref{cub-sen}. We also conduct class-conditional experiments on CUB and Oxford Flower. The generated samples of birds controlled by class labels are illustrated in Figure~\ref{cub-class}. The Inception Scores on CUB and Oxford Flower datasets are $6.43 \pm .09$ and $2.62 \pm .03$.

\begin{center}
\includegraphics[width= \textwidth]{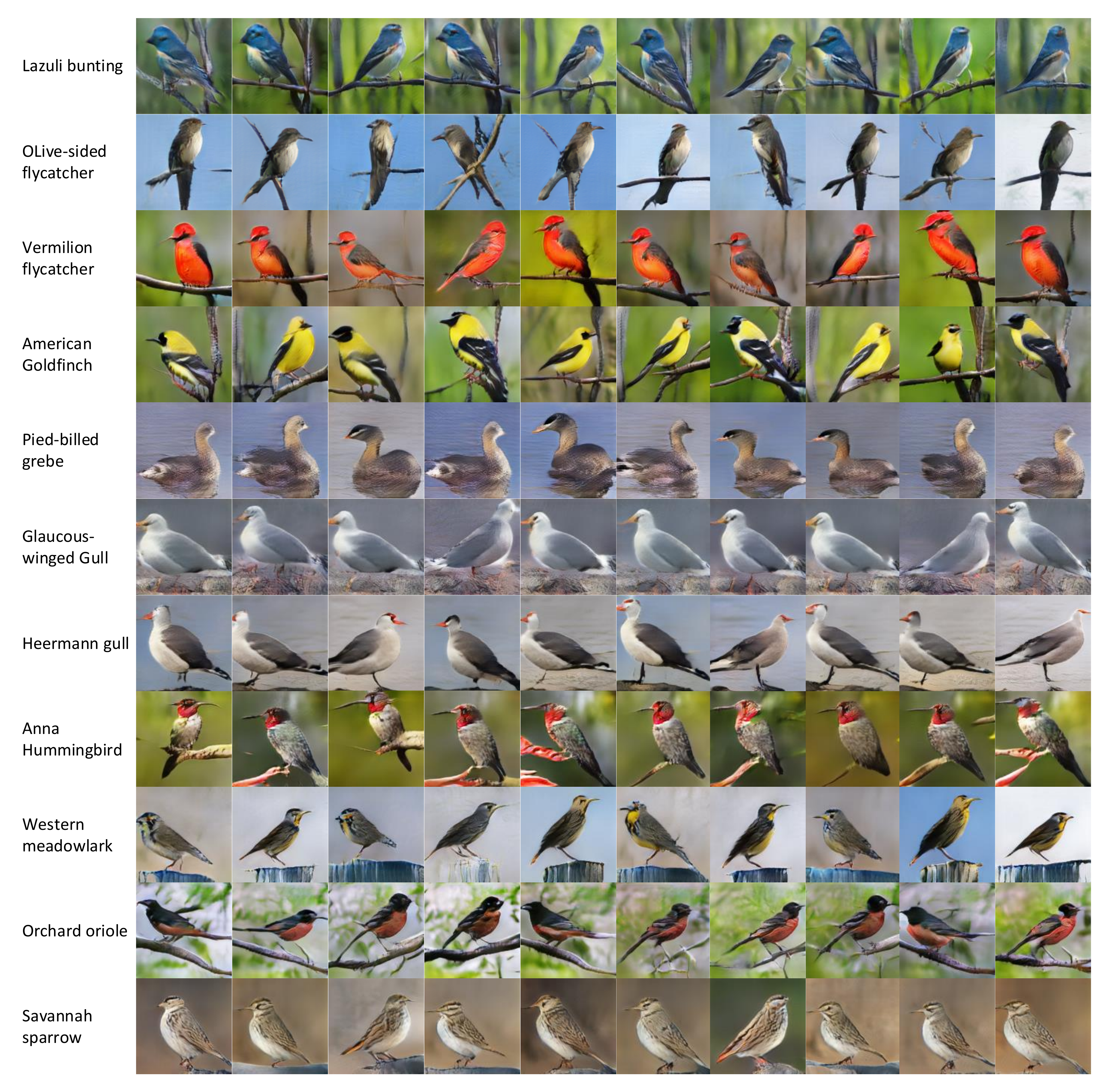}
\captionof{figure}{The generated 128 x 128 images controlled by class labels on CUB.}
\label{cub-class}
\end{center}

As shown in Figure \ref{cub-sen}, our model can achieve high resolution realistic images from text descriptions. The generated images have a fine-grained match with the descriptions such as the beak shape and the vivid parts. The Figure \ref{cub-class} also presents the high resolution class-conditional results, which show the realistic details (e.g., feather texture) and rich variety under the class labels. And the generated background is also clear and suitable.

\begin{figure}[!htbp]
\centering
\vspace*{-.3in}
\includegraphics[width= \textwidth]{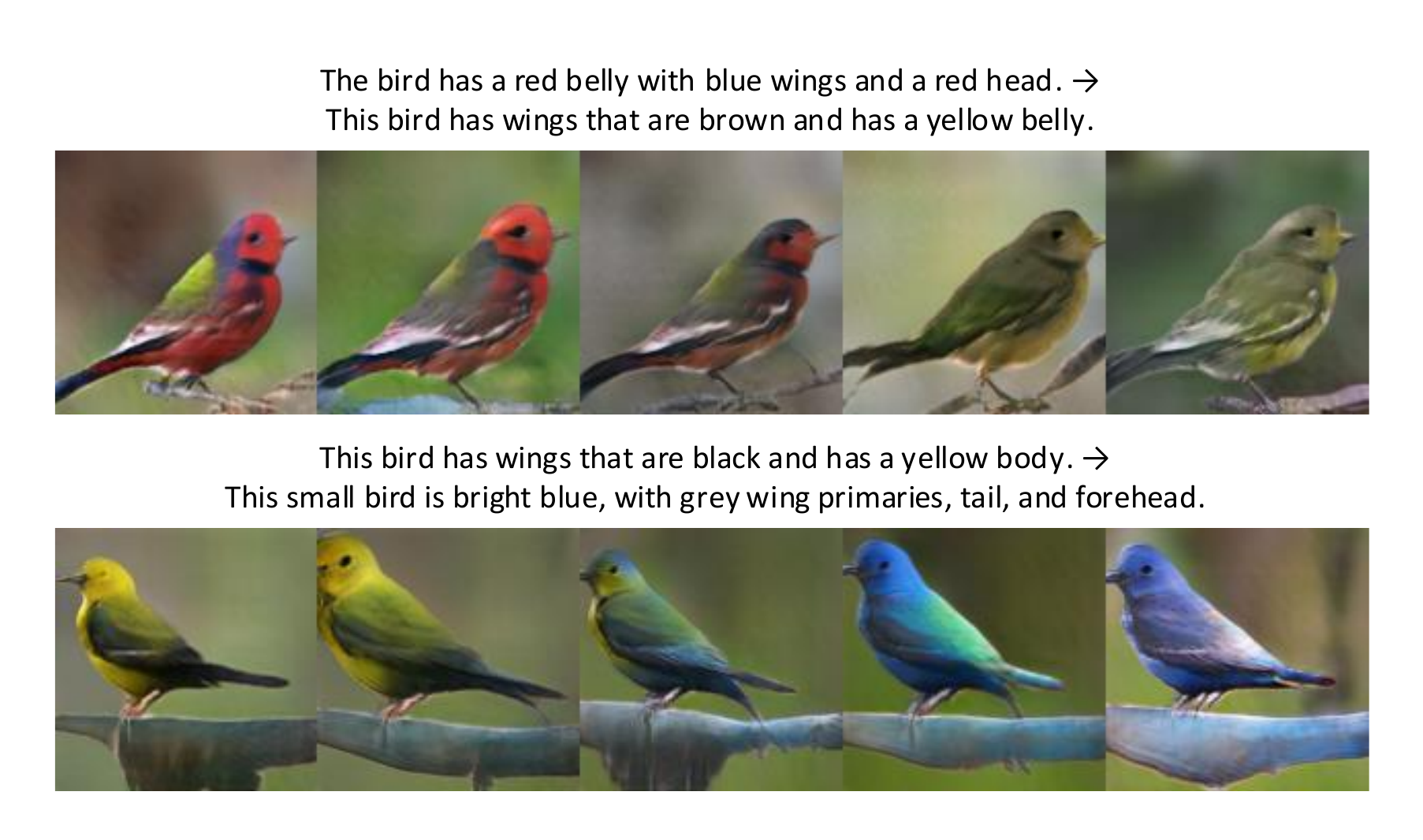}
\vspace*{-0.3in}
\caption{Images generated by interpolating between two sentence embeddings (Left to right). The noise is fixed for each row.}
\label{bird-interpolate}
\end{figure}

To further explore the smooth property of latent data manifold learned by our model, the linear interpolation results between two different text descriptions are shown in Figure \ref{bird-interpolate}. The noise is fixed so that the images variety is only influenced by the sentence semantics. We can see the interpolated images have a gradual change in the colors of different parts. It is also noted that the interpolated images have a tiny variety in the pose and background though the noise is fixed. We speculate that the learned latent distribution rather than a fixed representation introduce some disturbance to yield the richer variety.

\vspace*{-0.2in}
\section{Conclusion}
In this paper, we have proposed a variational generator framework for conditional GANs to catch semantic details behind the conditional input and the fine-grained images with rich diversity can be achieved. The variational inference depending on the conditional input is introduced into the generator to generate images from the inferred latent posterior. Experimental results show that the proposed model outperforms the state-of-the-art approaches on the class-conditional task and we also demonstrate the generalizability of the proposed framework in the more complex sentence-level task. The truncation technique is incorporated to further boost the generation performance. Furthermore, the modified auxiliary classifier can make training converge faster and weaken mode collapse. In the future work, we hope to generate images with multiple entities from sentence descriptions by introducing the knowledge of entity concepts.

\bibliography{vcgan}

\end{document}